\documentclass[pdflatex,sn-mathphys-num]{sn-jnl}

\usepackage{graphicx}%
\usepackage{multirow}%
\usepackage{amsmath,amssymb,amsfonts}%
\usepackage{amsthm}%
\usepackage{mathrsfs}%
\usepackage[title]{appendix}%
\usepackage{xcolor}%
\usepackage{textcomp}%
\usepackage{manyfoot}%
\usepackage{booktabs}%
\usepackage{algorithm}%
\usepackage{algorithmicx}%
\usepackage{algpseudocode}%
\usepackage{listings}%
\PassOptionsToPackage{hyphens}{url}
\usepackage{url}

\theoremstyle{thmstyleone}%
\theoremstyle{thmstyletwo}%

\theoremstyle{thmstylethree}%

\begin{document}

\title[Article Title]{An Efficient Classification Model for Cyber Text}


\author[1]{\fnm{Md Sakhawat} \sur{Hossen}}\email{sakhawat3003@gmail.com}

\author*[2]{\fnm{Md. Zashid Iqbal} \sur{Borshon}}\email{zashidiqbal1554@gmail.com}

\author[2]{\fnm{A. S. M.} \sur{Badrudduza}}\email{asmb.kanon@gmail.com}

\affil*[1]{\orgdiv{Department of EEE}, \orgname{Chittagong University of Engineering \& Technology}, \city{Chittagong}, \postcode{4349}, \country{Bangladesh}}

\affil[2]{\orgdiv{Department of ETE}, \orgname{Rajshahi University of Engineering \& Technology}, \city{Rajshahi}, \postcode{6204}, \country{Bangladesh}}


\abstract{The uprising of deep learning methodology and practice in recent years has brought about a severe consequence of increasing carbon footprint due to the insatiable demands on computational resources and power. The field of text analytics also experienced a massive transformation on this trend of monopolizing methodology. In this paper, the original TF-IDF algorithm has been modified and Clement Term Frequency–Inverse Document Frequency (CTF-IDF) has been proposed for data preprocessing. This paper primarily discusses on the effectiveness of classical machine learning techniques in text analytics with CTF-IDF and faster IRLBA algorithm for dimensionality reduction. The introduction of both of these techniques in the conventional text analytics pipeline ensures a more efficient, faster, and less computationally intensive application when compared with deep learning methodology regarding carbon footprint with minor compromise in accuracy. The experimental results also exhibit a manifold of reduction in time complexity and improvement of model accuracy for the classical machine learning methods discussed further in this paper.}


\maketitle

\section{Introduction}\label{sec1}

Since the advent of modern technology and the internet, the ubiquitous application of electronic media in academia and research, news publications, social media, government, and non-government sites has massively contributed to the upsurge of text data stored in digital appliances. To extract the essential information from this highly unstructured data, we must first employ a variety of data mining techniques to uncover potentially valuable patterns from this enormous amount of data. Text analytics is the process of retrieving unstructured data and transforming it into structured data with the application of suitable algorithms to find patterns and trends, and classify the texts into distinct groups \cite{mclaughlin2022textanalytics}. The standard text analytics methodology can be burdensome for any small machine when dealing with big unstructured data. Contemporary text classification task requires a copious amount of text documents in each training session. Consequentially, the feature space may explode with sparse and redundant data when transformed into a document frequency matrix. This ultimately results in a heavy toll on computational power and the time required to build any machine learning and deep learning model for the prediction and classification of text data. This is proverbially known as the curse of dimensionality. On the other hand, Deep Learning methods like Bidirectional LSTM and Transformer based models like Google’s BERT have shown significant improvements in the precision of text analysis but at a cost of huge computational time and resources, therefore aggravating the issue of carbon footprint. 

Term Frequency–Inverse Document Frequency (TF-IDF) is considered one of the stepping stones for transforming tokenized textual data. According to a 2015 survey, TF-IDF is used by 83\% of recommender systems based on textual data in digital libraries \cite{beel2016recommender}. This statistical metric quantifies the significance of a word in a corpus or collection of documents \cite{rajaraman_data}. The classical TF-IDF severely penalizes each word/token in the documents based on the frequency of the words among the whole corpus on a logarithmic scale which in return creates a wide range of TF-IDF values and sometimes diminishes the whole effect of various keywords \cite{cheng2018tfidf}. We have proposed a moderate and clement approach to address this issue with CTF-IDF. The experimental results showed improvement regarding model accuracy in implementing CTF-IDF over the classical TF-IDF on text classification tasks when combined with the proposed algorithm for dimensionality reduction. 

Dimensionality reduction is the method of converting high-dimensional data into a meaningful representation of lesser dimensionality \cite{maaten2009review}. This method is considered essential for transforming the features into a more compact form to increase the learning efficiency of the algorithms when the number of features exceeds significantly. Dimension reduction techniques can be utilized both with supervised and unsupervised methods. However, depending on the kind of method utilized, the properties of the dimensionality reduction technique change. For instance, dimensionality reduction techniques for unsupervised learning algorithms should work to reduce the loss of feature information. On the other hand, the goal should be to maximize class information in the case of supervised learning. There is no single strategy that works in every circumstance due to the complexity of the dimension reduction process. As a result, numerous dimension reduction techniques have been developed and proposed over the years and put to the test in various fields of study and application domains. 

In this paper, we adopted “The augmented implicitly restarted Lanczos bidiagonalization algorithm” for dimensionality reduction \cite{baglama2005augmented}. This algorithm computes partial singular values decomposition and finds a few of the largest or smallest singular values along with the singular vectors of a sparse or dense matrix. This is a fast and memory-efficient method that serves to alleviate the problem of employing complex machine-learning algorithms and improving the overall per-formance of the models at the same time. 

For the initial development of the methodology, a SPAM dataset was used that consists of 5000 text data collected primarily from phones as text messages \cite{almeida2011sms}. The dataset was classified into two basic categories: Spam and Ham (not Spam). After the development of the methodology on this dataset, a comparative analysis was done on another similar SMS Phishing dataset \cite{mishra2022sms} in terms of model accuracy and run time. For the analysis, classical machine learning techniques like Decision Tree and Support Vector Machine have been used to evaluate the robustness and efficiency of the proposed methodology in contrast with the traditional methodology in text analytics and deep learning model like Transformer (BERT).
\medskip
\\The main contributions of this paper are as follows:

\begin{itemize}
    \item  Introducing a modified data processing algorithm CTF IDF to reduce the penalty received by each term in the corpus.
    
    \item Incorporation of a faster and memory-efficient “Augmented implicitly restarted Lanczos bidiagonalization” algorithm for dimensionality reduction.
    
    \item Thecombined effect of both of these methods in the traditional text analytics pipeline expedited the computational time and reduced the requirement for computational resources.
    
    \item Interpretability of the Cybertexts based on features that influence the classification of texts as either spam or non spam (ham).
\end{itemize}

 The proposed method is efficient on addressing the issue of the rising carbon footprint due to the advent of Deep Learning methodology while still increasing the robustness
 of the trained models.

\section{Background Study}\label{sec2}
\subsection{Previous Works}\label{subsec1}
Xia Hu \textit{et al.} elaborately discussed the traditional methodology for text analytics, which consists of three key components: text preprocessing, text representation, and knowledge discovery \cite{cao2022rpnet}.

\begin{figure}[htbp]
\centering
\includegraphics[width=\textwidth]{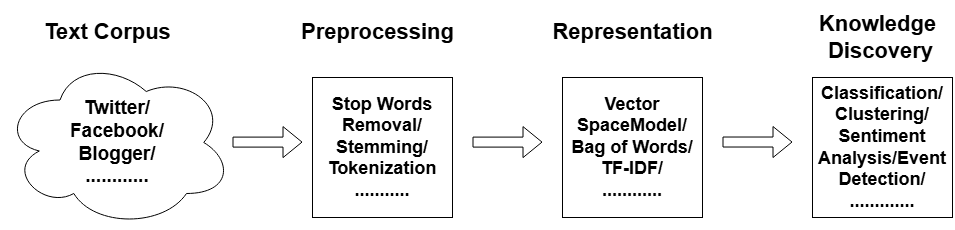}
\caption{Traditional framework for text analytics.}
\label{fig.1:traditional-framework}
\end{figure}

Our work mainly focuses on the representation stage with a new TF-IDF and a faster method for dimensionality reduction of the vector space. TF-IDF (Term Frequency-Inverse Document Frequency) is traditionally used as a statistical method for evaluating the importance of a word in a document in relation to a corpus of documents. TF-IDF has been widely used in text classification, such as spam filtering and sentiment analysis. Research has shown that TF-IDF often outperforms other feature selection methods, such as a bag of words and n-grams \cite{ahuja2019impact}. TF-IDF has also been used for keyword extraction, as it assigns high weight to important terms in a document and can identify the most relevant words for summarizing a document \cite{erra2015approximate}. In high-dimensional text data, TF-IDF is useful for dimensionality reduction, as it reduces the number of features while retaining important information \cite{dhar2018application}. In Recent research, TF-IDF has been combined with word embedding methods to improve the performance of text classification tasks \cite{deboom2016aggregation}. Hybrid Approaches like merging TF-IDF with other methods, such as word2vec, to improve its performance in text analysis \cite{liu2018improvedtfidf}. This paper introduces a modified TF-IDF for data representation. 

On the other hand, dimensionality reduction is an essential technique in text analytics for reducing the high dimensionality of textual data while retaining its most informative features. Principal Component Analysis (PCA) is a commonly used dimensionality reduction technique that involves projecting data onto a lower-dimensional space while retaining as much variance as possible. In text analytics, PCA has been used for tasks such as sentiment analysis, document classification, and topic modeling \cite{zu2003featuretransformation} \cite{han2004featuretransformation} \cite{zareapoor2015ieeb}. Latent Dirichlet Allocation (LDA) is a generative probabilistic model that discovers latent topics in a corpus of text. LDA has been used for tasks such as topic modeling, document classification, and information retrieval. \cite{blei2003lda} \cite{blei2001lda} \cite{jelodar2019lda}. Non-negative Matrix Factorization (NMF) is a matrix decomposition technique that factorizes a matrix into two non-negative matrices, which can be interpreted as representing latent topics and word distributions. NMF is suitable for tasks such as topic modeling, document clustering, and sentiment analysis \cite{nfmf_tool} \cite{lee1999nmf}. Singular Value Decomposition (SVD) is a matrix factorization technique used to decompose a matrix into its constituent parts. It is applied mostly in document classification, topic modeling, and information retrieval \cite{wang2011lsa}. For visualization of high-dimensional word embedding and document clustering t-distributed Stochastic Neighbor Embedding (t-SNE) is preferred \cite{liu2018visual} \cite{bamler2017dynamic} \cite{maaten2008tsne}. Word2Vec is frequently used for tasks such as text classification, sentiment analysis, and information retrieval \cite{ma2015word2vec} \cite{mikolov2013word2vec}. Another modern tool FastText has been applied in text classification, named entity recognition, and sentiment analysis \cite{joulin2016fasttext} \cite{almeida2017cnnfasttext}. GloVe is a technique that learns word embedding by factorizing a matrix of word co-occurrence statistics that has been implemented to find word similarity and text classification \cite{pennington2014glove}. 

Dimension reduction methods have been proven to be crucial for many text analytics tasks, and the choice of method depends on the specific task and the characteristics of the data. PCA, LDA, NMF, SVD, t-SNE, Word2Vec, FastText, and GloVe are some of the popular dimension reduction methods used in text analytics. In this paper, we experimented with a faster singular value decomposition method for dimensionality reduction.

\section{Proposed Method}\label{sec3}
The aim of this experiment is to provide an improved and efficient methodology in text analytics with classical machine learning algorithms. The basic framework consists of data preprocessing, feature extraction with Ctf-idf, projection of a Ctf-idf document vector into the SVD semantic space with IRLBA, and classification stages. The following subsection will provide a brief explanation of the success measure as well.

\begin{figure}[htbp]
\centering
\includegraphics[width=0.3\textwidth]{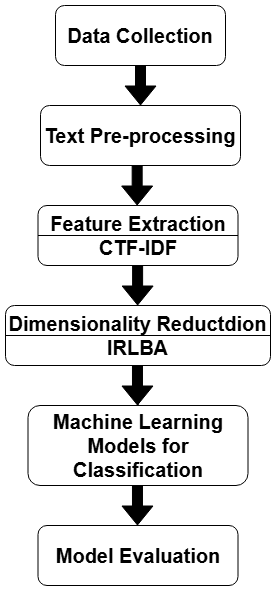}
\caption{Proposed method of the research.}
\label{fig.2:proposed-method}
\end{figure}

\subsection{Dataset Details}\label{subsec2}
A smaller and more compact SPAM dataset has been used for the development of the method, and the SMS Phishing dataset has been used as a benchmark dataset to check the robustness of the method.

\subsubsection{Spam Dataset}\label{subsubsec1}
The SMS Spam Collection consists of labeled SMS messages gathered for studying mobile phone spam \cite{almeida2011sms}. It is publicly available for research purposes. The dataset contains 5,574 authentic English text messages that are not encoded. These messages are categorized as either legitimate (ham) or spam. According to Collins Dictionary, spam messages are unsolicited electronic mail or messages sent simultaneously to a number of email addresses or mobile phones.

\subsubsection{SMS Phishing Dataset}\label{subsubsec2}
The dataset consists of categorized text messages gathered for SMS phishing investigation. It comprises 5,971 text messages labeled as either "Legitimate" (Ham), "Spam", or "Smishing". The collection encompasses 489 spam messages, 638 smishing messages, and 4,844 ham messages. For this research, the text documents labeled as ‘Smishing’ and ‘Spam’ are grouped together as ‘Spam’. This dataset contains raw message content that can be preprocessed for extracting further attributes to be followed by classification using Machine Learning techniques or can be used as labeled data in Deep Learning. Additionally, the dataset includes extracted attributes from malicious messages that aid in classifying messages as either malicious or legitimate.

\subsection{Preprocessing techniques}\label{subsec3}
The text documents will undergo pre-processing which involves various tasks such as tokenization, removal of stop words, conversion to lowercase, and stemming. Tokenization refers to dividing a text into tokens such as words or phrases. Stop words are words that appear frequently in texts, such as conjunctions or prepositions, regardless of the topic. Lowercase conversion involves changing all uppercase letters to lowercase letters before the classification stage. Stemming is the process of obtaining the root or stem of derived words, and the commonly used stemming process for English is Porter Stem, which was introduced by N. Milic-Frayling \cite{porter1997stemmer}.

\subsection{Feature Extraction}\label{subsec4}
\subsubsection{Document Frequency matrix}\label{subsubsec3}
Tokenization of the corpus is followed by the creation of a document frequency matrix to represent the connection between terms and documents. In this matrix, each row corresponds to a document, while each column corresponds to a term, and the value entered represents the frequency of the term’s occurrence within that particular document.

\subsubsection{N-gram modeling}\label{subsubsec4}
n-gram models are now widely applied in many computational fields including text analytics. There are many variants of n-grams depending on the sequential order. In order to reduce the load of quantitative analysis and sparsely distributed data, only the unigram model has been chosen to follow through the experiment.

\subsubsection{Modified TF-IDF (CTF-IDF)}\label{subsubsec5}
TF-IDF, short for Term Frequency-Inverse Document Frequency, combines two distinct measurements, TF and IDF, to analyze multiple documents. When dealing with multiple documents, TF-IDF is employed, leveraging the notion that uncommon and infrequent words provide greater insights into the content of a document compared to frequently occurring words across all documents. A modified CTF-IDF algorithm is proposed for this experiment. The modified algorithm assigns greater IDF values for the rarer terms in the whole corpus by calculating through the inverse hyperbolic sine function. The most infrequent terms in the corpus convey the most significance in classifying the document whereas the common terms should have very little significance in determining the nature of the document. In that case, CTF-IDF is more prominent in leveraging the rarity of any term and assigning much higher IDF value attached to it. CTF-IDF is also less inclement on penalizing the most frequent words so the CTF-IDF value of any word never diminishes. Not losing any information from the corpus is necessary on the application of matrix decomposition in the later steps during dimensionality reduction. The mathematical details are given below,

\begin{equation}
    tf(t, d) = \frac{f_d(t)}{\max\limits_{w \in d} f_d(w)}
\end{equation}

\begin{equation}
    idf(t, D) = \operatorname{arcsinh} \left( \frac{|D|}{|\{d \in D : t \in d\}|} \right)
\end{equation}

\begin{equation}
    tfidf(t, d, D) = tf(t, d) * idf(t, D)
\end{equation}

\begin{equation}
    tfidf'(t, d, D) = \frac{idf(t, D)}{|D|} + tfidf(t, d, D)
\end{equation}

Here,
\begin{itemize}
    \item $f_d(t)$ = Frequency of term $t$ in document $d$
    \item $D$ = Corpus of documents
\end{itemize}

\subsubsection{Dimensionality Reduction with IRLBA}\label{subsubsec6}
Another fundamental aspect of this experiment is to reduce the training time of each learning algorithm as much as possible while preserving accuracy. The proposed method of augmented implicitly restarted Lanczos bidiagonalization algorithm (IRLBA) is an extension of the Lanczos bidiagonalization algorithm that finds an estimated number of largest or the smallest singular values and corresponding singular vectors of a sparse or dense matrix using the mechanism of Baglama and Reichel \cite{baglama2005augmented}. It is a fast and memory-efficient method for truncated singular value decomposition and principal components analysis \cite{irlba2022}.

In this study, the transformation through IRLBA was iterated over many numbers of right singular vectors and an optimum 300 most significant right singular vectors have been chosen based on the descending order of singular values. The number of iterations determines the number of desired singular values to compute. The mathematical formulation is provided below. Given an input matrix $A$ of size $m \times n$, where $i = n$, this algorithm iteratively constructs two matrices $B$ and $C$, both bidiagonal, such that $B$ is similar to $A$.
\\

\begin{itemize}
    \item \textbf{Initialization}
    \begin{itemize}[label=\textbf{--}, left=0pt]
        \item Choose a starting vector $v_1$ of size $m \times 1$ with unit norm: $\|v_1\| = 1$
        \item Set $\beta_0 = 0$ and $v_0 = 0$
    \end{itemize}

    \item \textbf{Iteration} \\
    For $k = 1$ to $p$ (where $p$ is the desired number of singular values):
    \begin{itemize}[label=\textbf{--}, left=0pt]
        \item Compute $w_k = A \cdot v_k - \beta_{k-1} \cdot v_{k-1}$.
        \item $\alpha_k = \|w_k\|$.
        \item Normalize $w_k$: $v_{k+1} = \dfrac{w_k}{\alpha_k}$.
        \item Compute $z_k = A^T \cdot v_{k+1} - \alpha_k \cdot v_k$.
        \item $\beta_k = \|z_k\|$.
        \item Normalize $z_k$: $u_{k+1} = \dfrac{z_k}{\beta_k}$.
    \end{itemize}

    \item \textbf{Implicit Restart}
    \begin{itemize}[label=\textbf{--}, left=0pt]
        \item Compute the bidiagonalization of $B$ and $C$ for the first $p$ iterations using the Lanczos bidiagonalization algorithm.
    \end{itemize}

    \item \textbf{Augmentation}
    \begin{itemize}[label=\textbf{--}, left=0pt]
        \item Compute the singular value decomposition of the bidiagonal matrix $C$ of size $p \times p$: $C = U \cdot S \cdot V^T$.
    \end{itemize}

    \item \textbf{Implicit Restart (continued)}
    \begin{itemize}[label=\textbf{--}, left=0pt]
        \item Set $B = U^T \cdot B \cdot V$, which updates $B$ to be more similar to $A$.
        \item Repeat Steps 2–5 until convergence or desired accuracy is achieved.
    \end{itemize}
\end{itemize}

At the end of the algorithm, B and C will be similar, and the singular values of A can be computed from the singular values of C.

\subsection{Classification Methods}\label{subsec5}
For simplicity, Support Vector Machine and Decision Tree Classifiers have been chosen as the classical machine learning algorithms for classification in this experiment. On the other hand, the deep learning-based Transformer model BERT has been employed for comparative purposes with the proposed methodology. 

SVM is a learning algorithm designed for solving two-group classification problems, as originally introduced by \cite{cortes1995svm}. In this case, SVM is employed to categorize texts into positive or negative classes. SVM is particularly effective for text classification due to its ability to handle a large number of features with few examples when the problems can be linearly separated, as stated in \cite{joachims1998text}. 

Decision tree classifiers are widely recognized as one of the most popular and prominent approaches for representing classifiers in data classification. Decision Trees often replicate human cognitive processes when making decisions, thereby making them easily comprehensible and interpretable. A 10- fold cross-validation method has been employed during the training session. 

Bidirectional Encoder Representations from Transformers (BERT) is a highly sought-after new language representation model designed to pre-train deep bidirectional representations from the unlabeled text by joint conditioning on both left and right contexts in all layers \cite{devlin2018bert}. In this experiment, Multilingual BERT (mBERT) is used and it is flexible in providing sentence representations for 104 languages \cite{libovicky2019language}. It is recommended that no data preprocessing is required for modeling in BERT.

\subsection{Performance Parameters}\label{subsec6}
Four effective measures that have been used in this study are based on confusion matrix output, which are Sensitivity, Specificity, Balanced Accuracy, and Training time.

\begin{itemize}
    \item Sensitivity or Recall (True Positive Rate) = TP / (TP + FN)
    \item Positive Predictive Value or Precision = TP / (TP + FP)
    \item F-Measure = $2 \cdot \dfrac{\text{Precision} \cdot \text{Recall}}{\text{Precision} + \text{Recall}}$
    \item Training Time = Amount of time required to train the Learning Algorithm
\end{itemize}

The usefulness of these metrics is ubiquitous in text classification for comparative analysis among numerous learning algorithms. The F-measure serves as a middle ground between recall and precision, representing a balance between the two. Its value is significant only when both recall and precision are at high levels. When $\alpha$ (a parameter) equals $0$, the F-measure is equivalent to recall, while $\alpha = 1$ makes it equivalent to precision. The F-measure ranges from 0 to 1, with 0 indicating that no relevant documents were retrieved and 1 indicating that all retrieved documents are relevant and all relevant documents were retrieved.

\subsection{Interpretability}\label{subsec7}
For the interpretability of the models, Decision trees have been used primarily that can provide valuable insights into how the model classifies texts as spam or non-spam. The decision tree for both the SPAM data and the SMS Phishing data has been analyzed carefully to understand how it separates spam from non-spam text. Each internal node in the tree represents a decision based on a particular feature, and each leaf node represents a final decision (spam or non-spam). Attention has been paid to the split points and the features used for splitting, as they indicate the most important attributes for the classification. Features higher up in the tree that appears closer to the root node have been identified as the most important features, as they are the words or word stem that strongly influence the spam classification.

\subsection{Execution Environments}\label{subsec8}
All the modeling with classical learning algorithms and analyses were carried out in R (v.~4.0.3) on an old computer equipped with a Core 2 Duo processor (Operating at 3.00 GHz base) and 8~GB of RAM. The modeling with Transformers (mBERT) was executed in a computer with an Intel Core i5-7500 processor (Base clock 3.40 GHz), 32~GB of RAM, and NVIDIA GeForce GTX 1050Ti graphics card with 4~GB DDR5 RAM and 768 CUDA cores. Text preprocessing, Decision Trees, SVMs, IRLBA, and Transformers were respectively implemented using the \texttt{quanteda}, \texttt{caret}, \texttt{e1071}, \texttt{irlba}, \texttt{TensorFlow}, \texttt{reticulate}, and \texttt{keras} packages in R.

\section{Results and Discussions}\label{sec4}
For each trained model, a couple of comparisons were made in terms of model accuracy and training time. Each dataset was split into two parts, one for training and the other for testing: SPAM dataset with a ratio of 70:30: 70\% for training and 30\% testing. The SMS Phishing dataset was also partitioned with a ratio of 70:30. Table I, 2, 3, and 4 summarizes the performance metrics for the decision tree and Support Vector Machine Model for the SPAM and SMS Phishing dataset respectively. Table 5 accumulates the results after training both of the SPAM and SMS Phishing data with the Transformers model (mBERT). 

In the first phase for the SPAM dataset, all the raw text data went through the preprocessing stages followed by feature extraction through traditional tf-idf and modified tf-idf (Ctf-idf) respectively. Decision tree models were then built upon the tf-idf and Ctf-idf transformed data for classification. From the table I, it can be seen that there are no discernible changes in the performance metrics for both cases in the decision tree models. The time required to train each of the models was 13 minutes and 12 minutes respectively.

In the second phase for the SPAM dataset, the tf-idf and Ctf-idf feature extracted models were transformed by IRLBA.

\begin{table}[htbp]
\caption{Spam data 10-fold cross-validation performance metrics for Decision Tree model.}
\label{tab:spam-performance}
\centering
\begin{tabular}{@{}lcccc@{}}
\toprule
Models              & Precision & Recall  & F1-score & Training time \\
\midrule
TF-IDF              & 0.9627    & 0.9687  & 0.9657   & 13~min        \\
CTF-IDF             & 0.9627    & 0.9687  & 0.9657   & 12~min        \\
TF-IDF (IRLBA)      & 0.9869    & 0.9420  & 0.9640   & 17~sec        \\
CTF-IDF (IRLBA)     & 0.9889    & 0.9508  & 0.9700   & 16~sec        \\
\bottomrule
\end{tabular}
\end{table}

After the transformation, it required only 17 seconds to train a decision tree model on tf-idf and 16 seconds on Ctf-idf transformed corpus (Table 1). The computational time was
significantly reduced by the application of IRLBA. Table 1 also shows the combined effect of Ctf-idf and IRLBA transformation helps to increase the F1-score from 96.57\% to
97\% in contrast with the tf-idf and IRLBA transformed data where it decreases slightly

\begin{table}[htbp]
\caption{Spam data performance metrics for SVM model.}
\label{tab:spam-svm}
\centering
\begin{tabular}{@{}lcccc@{}}
\toprule
Models             & Precision & Recall  & F1-score & Training time \\
\midrule
TF-IDF             & 0.9807    & 0.9827  & 0.9817   & 35~sec        \\
CTF-IDF            & 0.9793    & 0.9832  & 0.9812   & 37~sec        \\
TF-IDF (IRLBA)     & 0.9750    & 0.9912  & 0.9830   & 5~sec         \\
CTF-IDF (IRLBA)    & 0.9783    & 0.9965  & 0.9873   & 5~sec         \\
\bottomrule
\end{tabular}
\end{table}

All the procedures were the same for Support Vector Machine as well. It can be seen from table 2 that F1-Scores are hovering over 98\% for preliminary tf-idf and Ctf-idf transformed data with 35 and 37 seconds of training time for each model respectively. After the projection of tf-idf and Ctf-idf transformed data in the semantic space
through IRLBA, the overall performance of the models increased. It can be seen from Table 2 that the F-Scores improved in both cases but the Ctf-idf-IRLBA transformed
data performed better with an F-Score of 98.73\%. Not only that, the computational time was also vastly reduced to 5 seconds only for training each of the models in SVM.

For SPAM dataset, the Transformers (mBERT) models raised the training and the validation accuracy slightly above 99\% but with the expense of a huge computational power (Table
5). It required NVIDIA graphics card to run all the CUDA cores simultaneously for up to 1:30 hours on the minimum level to train the model.

\begin{table}[htbp]
\caption{SMS phishing data 10-fold cross-validation performance metrics for Decision Tree model.}
\label{tab:sms-dt}
\centering
\begin{tabular}{@{}lcccc@{}}
\toprule
Models             & Precision & Recall  & F1-score & Training time \\
\midrule
TF-IDF             & 0.9767    & 0.9184  & 0.9467   & 10.33~min     \\
CTF-IDF            & 0.9767    & 0.9184  & 0.9467   & 9.67~min      \\
TF-IDF (IRLBA)     & 0.9413    & 0.9601  & 0.9506   & 22.09~sec     \\
CTF-IDF (IRLBA)    & 0.9520    & 0.9683  & 0.9601   & 19.46~sec     \\
\bottomrule
\end{tabular}
\end{table}

The robustness of the methodology is tested on the SMS Phishing dataset. Table 3 shows, without the application of IRLBA, the Ctf-idf performs no better than the tf-idf
transformed data with an F1-score of 94.67\% in decision tree models. But, the training time for the Ctf-idf model was slightly better than the tf-idf model taking 9.67 minutes.
After the IRLBA transformation of the tf-idf and Ctf-idf models, the F-1 scores increased above 95\% for both models but again Ctf-idf-IRLBA performed better. Table 3 also shows the tremendous reduction in training time requiring approximately 20 seconds which took almost 10 minutes previously.

\begin{table}[htbp]
\caption{SMS phishing data performance metrics for SVM model.}
\label{tab:sms-svm}
\centering
\begin{tabular}{@{}lcccc@{}}
\toprule
Models             & Precision & Recall  & F1-score & Training time \\
\midrule
TF-IDF             & 0.9832    & 0.9642  & 0.9736   & 14.31~sec     \\
CTF-IDF            & 0.9854    & 0.9738  & 0.9796   & 12.90~sec     \\
TF-IDF (IRLBA)     & 0.9703    & 0.9913  & 0.9807   & 9.82~sec      \\
CTF-IDF (IRLBA)    & 0.9737    & 0.9945  & 0.9840   & 9.80~sec      \\
\bottomrule
\end{tabular}
\end{table}

For Support Vector Machine, Table 4 shows the methodology exhibits the same characteristic reduction in training time but the difference is not that significant.
Again, the Ctf-idf transformed model worked better on SMS Phishing data than the tf-idf model before transforming the models through IRLBA. After the application of IRLBA it
can be seen from Table 4 that the F1-Scores again improved up to 98\% for both of the models and still Ctf-idf-IRLBA maintained better performance.

\begin{table}[htbp]
\caption{Transformer (mBERT) Performance Metrics for SPAM and SMS Phishing Data.}
\begin{tabular}{|l|l|l|l|l|}
\hline
Data                                                         & \begin{tabular}[c]{@{}l@{}}No. \\ of Epoch\end{tabular} & \begin{tabular}[c]{@{}l@{}}Training \\ Accuracy\end{tabular} & \begin{tabular}[c]{@{}l@{}}Validation \\ Accuracy\end{tabular} & \begin{tabular}[c]{@{}l@{}}Training \\ Time\end{tabular} \\ \hline
SPAM Data                                                    & 7                                                       & 0.9913                                                       & 0.9904                                                         & 1:30 hrs                                                 \\ \hline
\begin{tabular}[c]{@{}l@{}}SMS Phishing \\ Data\end{tabular} & 15                                                      & 0.9976                                                       & 0.9888                                                         & 1:43 hrs                                                 \\ \hline
\end{tabular}
\end{table}

In the case of the Transformers (mBERT) model, table 5 shows the model performs a little better than the SVM with a validation accuracy of 98.88\% for the SMS Phishing dataset. As expected the transformer model took a considerable amount of training time over 1 hour and 35 minutes with the help of a NVIDIA graphics card at the backend with all the CUDA cores running simultaneously. 

For interpretability of the spam texts in the SPAM dataset, the following stemmed words have been found influential according to the importance of making decisions on the decision tree model: call, txt, prize, guarantee, claim, won, urgent, tone, service. The important features or stemmed words for the interpretability of the spam texts in the SMS Phishing dataset are as follows: opinion, jada, spl, stylish, online, pls, want, silent, simple, character, loveable, reply, matur, come, free, can, talk, sometimes, working, overtime, sir, sent, email, log, payment, portal, send, tax, pay, citizen. The presence of these words or the word stem is suggestive of the texts being considered spam. 

It is quite evident from the experiment that Ctf-idf transformation of the dataset with the combination of IRLBA algorithm for dimensionality reduction is significantly faster in training any classical machine learning model and at the same time it improves the performance of the model after shrinking and transforming the feature space from thousands of columns to a handful of informative columns. The transformers (BERT)
models are great in producing state-of-the-art accuracy but with a huge cost of computational power and producing a higher carbon footprint.

\section{Conclusions and Future Work}\label{sec5}
This study presents a comprehensive approach to tackle the computational and environmental challenges associated with text analytics, specifically in the context of Spam detection. It introduces the innovative CTF-IDF method and IRLBA for dimensionality reduction, significantly reducing the need for computational resources and improving model accuracy at the same time. 

The utilization of CTF-IDF, a novel approach to transforming tokenized text data, addressed the issue of severe penalties imparted by traditional TF-IDF method, often resulting in the loss of the effects of various keywords. Our results illustrated that the deployment of CTF-IDF enhances model accuracy in text classification tasks. The incorporation of the IRLBA algorithm further contributed to the efficiency and speed of the proposed methodology. 

Through our experimentation with both the SPAM dataset and the SMS Phishing dataset, the models exhibited notable improvement in computational time and reduced requirements for computational resources. Moreover, they showed competitive accuracy rates when compared with sophisticated deep learning models like BERT, thus providing a viable alternative. Looking forward, our research will focus on enhancing the proposed methodology. For CTF-IDF, we plan to develop an adaptive model that can adjust its penalty metrics based on the specifics of the text corpus. As for the IRLBA algorithm, we aim to optimize its performance further for both sparse and dense matrix types, and potentially explore its application in other areas of data analytics. 

Moreover, we will investigate the generalizability of our proposed methodology in different domains and diverse datasets. It would be intriguing to examine the efficiency of our approach in other text analytics tasks like sentiment analysis, topic modeling, or document clustering.

Finally, the intersection of machine learning and environmental sustainability remains a compelling field for future exploration. The current study focuses on reducing the computational resources in text analytics, but the concept can be extended to other machine learning tasks. The quest to make AI ”greener” should continue with more research on energy-efficient algorithms and models, not only for the sake of our environment but also to ensure the sustainable development of machine learning technologies. The research community must strive for a balance between algorithmic innovation and computational responsibility. In this context, our work serves as an important step towards shaping the future of responsible AI. 

\section*{Acknowledgment}\label{sec6}
The authors received no external funding for this research.

\bibliography{ZahuBiblio}

\end{document}